\documentclass{article}

\usepackage[english]{babel}
\usepackage[utf8x]{inputenc}
\usepackage{amsmath}
\usepackage{float}
\usepackage{graphicx}
\usepackage{wrapfig}
\usepackage[colorinlistoftodos]{todonotes}

\usepackage{times}
\usepackage{graphicx} % more modern
\usepackage{subfigure} 

\usepackage{natbib}

\usepackage{amsmath}
\usepackage{bm}

\usepackage{hyperref}

\usepackage[accepted]{icml2016} 

\icmltitlerunning{This paper describes an older version of DeepLIFT; see \url{https://arxiv.org/abs/1704.02685} for the new version}

\begin{document} 

\twocolumn[
\icmltitle{Not Just A Black Box: \\ 
           Learning Important Features Through Propagating Activation Differences}

% unless you've provided the [accepted] option to the icml2015
\begin{center}{ {\bf Avanti Shrikumar$^1$} (avanti@stanford.edu), {\bf Peyton Greenside$^2$} (pgreens@stanford.edu)\\{\bf Anna Y. Shcherbina$^2$} (annashch@stanford.edu), {\bf Anshul Kundaje$^{1,3}$} (akundaje@stanford.edu)}\end{center}
\icmladdress{1. Department of Computer Science, Stanford University, CA, USA\\2. Biomedical Informatics, Stanford University, CA, USA\\3. Department of Genetics, Stanford University, CA, USA}

\icmlkeywords{DeepLIFT, DeepLIFT Technologies, LIFTPAD, deep learning, interpretable deep learning, interpretability, interpretable, interpretation, visualization, visualizing, saliency, saliency map, importance, pixel-wise, neural networks, feature importance}

\vskip 0.3in
]

\begin{abstract} 
{\bf This paper describes an older version of DeepLIFT. See \url{https://arxiv.org/abs/1704.02685} for the new version}. The purported ``black box'' nature of neural networks is a barrier to adoption in applications where interpretability is essential. Here we present DeepLIFT (Learning Important FeaTures), an efficient and effective method for computing importance scores in a neural network. DeepLIFT compares the activation of each neuron to its `reference activation' and assigns contribution scores according to the difference. We apply DeepLIFT to models trained on natural images and genomic data, and show significant advantages over gradient-based methods.
\end{abstract} 

\section{Introduction}
\label{introduction}

As neural networks become increasingly popular, their ``black box" reputation is a barrier to adoption when interpretability is paramount. Understanding the features that lead to a particular output builds trust with users and can lead to novel scientific discoveries. \citet{Simonyan2013-hk} proposed using gradients to generate saliency maps and showed that this is closely related to the deconvolutional nets of \citet{Zeiler2014-sk}. Guided backpropagation \citep{Springenberg2014-gg} is another variant which only considers gradients that have positive error signal. As shown in Figure 2, saliency maps can be substantially improved by simply multiplying the gradient with the input signal, which corresponds to a first-order Taylor approximation of how the output would change if the input were set to zero; as we show, the layer-wise relevance propagation rules described in \citet{Bach2015-pq} reduce to this approach, assuming bias terms are included in the denominators.

Gradient-based approaches are problematic because activation functions such as Rectified Linear Units (ReLUs) have a gradient of zero when they are not firing, and yet a ReLU that does not fire can still carry information (Figure 1). Similarly, sigmoid or tanh activations are popular choices for the activation functions of gates in memory units of recurrent neural networks such as GRUs and LSTMs \citep{Chung2014-uj, Hochreiter1997-yn}, but these activations have a near-zero gradient at high or low inputs even though such inputs can be very significant.

\begin{figure}[!ht]
\vspace{-5px}
\begin{center}
\includegraphics[scale=0.5]{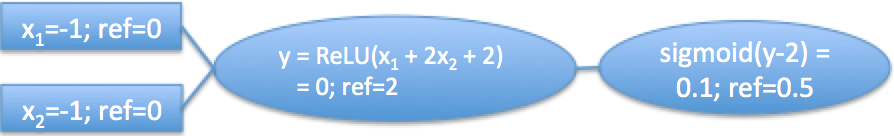}
\caption{Simple network with inputs $x_1$ and $x_2$ that have reference values of 0. When $x_1 = x_2 = -1$, output is 0.1 but the gradients w.r.t $x_1$ and $x_2$ are 0 due to inactive ReLU $y$ (which has activation of $2$ under reference input). By comparing activations to their reference values, DeepLIFT assigns contributions to the output of $\left((0.1-0.5)\frac{1}{3}\right)$ to $x_1$ and $\left((0.1-0.5)\frac{2}{3}\right)$ to $x_2$.}
\end{center}
\vspace{-10px}
\end{figure}

We present DeepLIFT, a method for assigning feature importance that compares a neuron's activation to its `reference', where the reference is the activation that the neuron has when the network is provided a `reference input' (the reference input is defined according to what is appropriate for the task at hand). This addresses the limitation of gradient-based approaches because the difference from the reference may be non-zero even when the gradient is zero.

\vspace{-10px}
\section{DeepLIFT Method}
\label{DeepLIFT}

We denote the contribution of neuron $x$ to neuron $y$ as $C_{xy}$. Let the activation of a neuron $n$ be denoted as $A_n$. Further, let the \emph{reference} activation of neuron $n$ be denoted $A_n^0$, and let the $A_n - A_n^0$ be denoted as $\delta_n$. We define our contributions $C_{xy}$ to satisfy the following properties.

\vspace{-5px}
\subsection{Summation to $\delta$}

For any set of neurons $S$ whose activations are minimally sufficient to compute the activation of $y$ (that is, if we know the activations of $S$, we can compute the activation of $y$, and there is no set $S' \subset S$ such that $S'$ is sufficient to compute the activation of $y$ - in layman's terms, $S$ is a full set of non-redundant inputs to $y$), the following property holds:
\begin{equation}
\sum_{s \in S} C_{sy} = \delta_y
\end{equation}
That is, the sum over all the contributions of neurons in $S$ to $y$ equals the difference-from-reference of $y$.

\subsection{Linear composition}

Let $O_x$ represent the output neurons of $x$. The following property holds:\\
\begin{equation}
C_{xy} = \sum_{o \in O_x} \frac{C_{xo}}{\delta_o}C_{oy}
\end{equation}
In layman's terms, each neuron `inherits' a contribution through its outputs in proportion to how much that neuron contributes to the difference-from-reference of the output.

\subsection{Backpropagation Rules}

We show that the contributions as defined above can be computed using the following rules (which can be implemented to run on a GPU). The computation is reminiscent of the chain rule used during gradient backpropagation, as equation $2$ makes it possible to start with contribution scores of later layers and use them to find the contribution scores of preceding layers. To avoid issues of numerical stability when $\delta_n$ for a particular neuron is small, rather than computing the contribution scores explicitly, we instead compute \emph{multipliers} $m_{xy}$ that, when multiplied with the difference-from-reference, give the contribution:
\begin{equation}
m_{xy} \delta_x = C_{xy}
\end{equation} 
Let $t$ represent the target neuron that we intend to compute contributions to, and let $O_x$ represent the set of outputs of $x$. We show that:
\begin{equation}
m_{xt} = \sum_{y \in O_x} m_{xy}m_{yt}
\end{equation} 
The equation above follows from the linear composition property and the definition of the multipliers, as proved below:
\begin{equation}
\begin{aligned}
C_{xt} &= \sum_{y \in O_x} \frac{C_{xy}}{\delta_y}C_{yt}\\
m_{xt} \delta_x &= \sum_{y \in O_x} \frac{C_{xy}}{\delta_y}(m_{yt} \delta_y) = \sum_{y \in O_x} C_{xy}m_{yt}\\
m_{xt} &= \sum_{y \in O_x} \frac{C_{xy}}{\delta_x}m_{yt} = \sum_{y \in O_x} m_{xy} m_{yt}
\end{aligned}
\end{equation}

In the equations below, $I_y$ denotes the set of inputs of $y$.

\subsubsection{Affine functions}
Let
\begin{equation}
A_y = \left(\sum_{x \in I_y} w_{xy} A_x\right) + b  
\end{equation}
Then $m_{xy} = w_{xy}$

{\bf Proof.} We show that $\delta_y = \sum_{x \in I_y} m_{xy} \delta_x$.

Using the fact that $A_n = A_n^0 + \delta_n$, we have:
\begin{equation}
\begin{aligned}
(A_y^0 + \delta_y) &= \left(\sum_{x \in I_y} w_{xy} (A_x^0 + \delta_x) \right) + b \\
                              &= \left(\sum_{x \in I_y} w_{xy} A_x^0 \right) + b + \sum_{x \in I_y} w_{xy} \delta_x
\end{aligned}
\end{equation} 
We also note that the reference activation $A_y^0$ can be found as follows:
\begin{equation}
A_y^0 = \left(\sum_{x \in I_y} w_{xy} A_x^0 \right) + b
\end{equation}
Thus, canceling out $A_y^0$ yields:
\begin{equation}
\begin{aligned}
\delta_y &= \sum_{x \in I_y} w_{xy} \delta_x = \sum_{x \in I_y} m_{xy} \delta_x
\end{aligned}
\end{equation} 

%\subsubsection{Maxpooling over convolutional filters}

%We consider the case of maxpooling over a set of neurons in a convolutional layer which all belong to the same convolutional filter:
%\begin{equation}
%A_y = \max_{x \in I_y} A_x
%\end{equation}
%We make the assumption that all instances of a single convolutional filter have the same reference activation, i.e. $A_x^0$ is the same for all $x$ in $I_y$ (this would not hold if the chosen reference input were not uniform across all positions; the appropriate multiplier in that situation can be computed using the general formula for $m_{xy}$ provided above). Then we have:
%\begin{equation}
%m_{xy} = \bm{1}\{A_x = A_y\}
%\end{equation}
%Where $\bm{1}\{\}$ is the indicator function. If a symbolic computation package is used, then the gradient of $y$ with respect to $x$ can be used in place of $\bm{1}\{A_x = A_y\}$.

%{\bf Proof.} Because $A_x^0$ is assumed to be the same for all $x$, we have that $A_y^0 = A_x^0$. As $A_y = A_y^0 + \delta_y$, we get:
%\begin{equation}
%\begin{aligned}
%\delta_y &= A_y - A_y^0 = A_y - A_x^0\\
%             &= \bm{1}\{A_x = A_y\} (A_x - A_x^0)\\
%             &= \bm{1}\{A_x = A_y\} \delta_x = m_{xy} \delta_x
%\end{aligned}
%\end{equation}

\subsubsection{Max operation}

We consider the case of max operation such as a maxpool:
\begin{equation}
A_y = \max_{x \in I_y} A_x
\end{equation}
Then we have:
\begin{equation}
m_{xy} = \bm{1}\{A_x = A_y\}\frac{\delta_y}{\delta_x}
\end{equation}
Where $\bm{1}\{\}$ is the indicator function. If a symbolic computation package is used, then the gradient of $y$ with respect to $x$ can be used in place of $\bm{1}\{A_x = A_y\}$.

{\bf Proof.}
\begin{equation}
\begin{aligned}
\sum_{x \in y} m_{xy} \delta_x &= \left(\sum_{x \in y} \bm{1}\{A_x = A_y\}\frac{\delta_y}{\delta_x}\right)\delta_x\\ &= \sum_{x \in y} \bm{1}\{A_x = A_y\}\delta_y = \delta_y
\end{aligned}
\end{equation}

\subsubsection{Maxout units}
A maxout function has the form
\begin{equation}
A_y = \max_{i = 1}^n \left(\sum_x w_{xy}^i A_x\right) + b^i
\end{equation}
i.e. it is the max over $n$ affine functions of the input vector $\vec{x}$. For a given vector of activations $A_{\vec{x}}$ of the inputs, we split $A_{\vec{x}} - A_{\vec{x}}^0$ into segments such that over each segment $s$, a unique affine function dominates the maxout and the coefficient of an individual input $x$ over that segment is $w(s)_{xy}$. Let $l(s)$ denote the fraction of $A_{\vec{x}} - A_{\vec{x}}^0$ in segment $s$. We have:
\begin{equation}
m_{xy} = \sum_s l(s)w(s)_{xy}
\end{equation}
Intuitively speaking, we simply split the piecewise-linear maxout function into regions where it is linear, and do a weighted sum of the coefficients of $x$ in each region according to how much of $A_{\vec{x}} - A_{\vec{x}}^0$ falls in that region.

\subsubsection{Other activations}
The following choice for $m_{xy}$, which is the same for all inputs to $y$, satisfies summation-to-delta:\\
\begin{equation}
m_{xy} = \frac{\delta_{y}}{\sum_{x' \in I_y} \delta_{x'}}
\end{equation}
This rule may be used for nonlinearities like ReLUs, PReLUs, sigmoid and tanh (where $y$ has only one input). Situations where the denominator is near zero can be handled by applying L'hopital's rule, because by definition:
\begin{equation}
\delta_{y} \rightarrow 0 \text{ as} \sum_{x \in I_y} \delta_{x} \rightarrow 0
\end{equation}

\subsubsection{Element-wise products}
Consider the function:
\begin{equation}
A_y = A_y^0 + \delta_y = (A_{x_1}^0 + \delta_{x_1}) (A_{x_2}^0 + \delta_{x_2})
\end{equation}
We have:\\
\begin{equation}
\begin{aligned}
\delta_y &= (A_{x_1}^0 + \delta_{x_1})(A_{x_2}^0 + \delta_{x_2}) - (A_{x_1}^0A_{x_2}^0)\\
         &= A_{x_1}^0\delta_{x_2} + A_{x_2}^0\delta_{x_1} + \delta_{x1}\delta_{x_2}\\
         &= \delta_{x_1}\left(A_{x_2}^0 + \frac{\delta_{x_2}}{2}\right) + \delta_{x_2}\left(A_{x_1}^0 + \frac{\delta_{x_1}}{2}\right)
\end{aligned}
\end{equation}
Thus, viable choices for the multipliers are $m_{x_1 y} = A_{x_2}^0 + 0.5\delta_{x_2}$ and $m_{x_2 y} = A_{x_1}^0 + 0.5\delta_{x_1}$

%\subsubsection{Individual nonlinearity}

%Here we describe the rule for nonlinear transformations such as the ReLU, sigmoid and tanh which are applied to individual inputs, typically following some affine transformation. We have:
%\begin{equation}
%A_y = f(A_x)
%\end{equation}
%Where $f$ is the nonlinear transformation. When $\delta_x$ is large in magnitude, we can compute $m_{xy}$ according to its definition as follows:
%\begin{equation}
%m_{xy} = \frac{\delta_y}{\delta_x}
%\end{equation}
%When $\delta_x$ is small, the term on the right approaches $\frac{0}{0}$. However, as $\delta_y \rightarrow 0$ as $\delta_x \rightarrow 0$, we have:
%\begin{equation}
%\lim_{\delta_x \rightarrow 0} \frac{\delta_y}{\delta_x} = \frac{d \delta_y}{d \delta_x}
%\end{equation}
%Thus, when $\delta_x$ is close to zero, $m_{xy}$ is simply the derivative of $\delta_y$ with respect to $\delta_x$. Because $\delta_n = A_n - A_n^0$ and $A_n^0$ is a constant, this is the same as $\frac{dA_y}{dA_x}$ when $\delta_x$ is near $0$.

\subsection{A note on final activation layers}

Activation functions such as a softmax or a sigmoid have a maximum $\delta$ of 1.0. Due to the \emph{summation to $\delta$} property, the contribution scores for individual features are lower when there are several redundant features present. As an example, consider $A_t = \sigma(A_y)$ (where $sigma$ is the sigmoid transformation) and $A_y = A_{x_1} + A_{x_2}$. Let the default activations of the inputs be $A_{x_1}^0 = A_{x_2}^0 = 0$. When $x_1 = 100$ and $x_2 = 0$, we have $C_{x_1 t} = 0.5$. However, when both $x_1 = 100$ and $x_2 = 100$, we have $C_{x_1 t} = C_{x_2 t} = 0.25$. To avoid this attenuation of contribution in the presence of redundant inputs, we can use the contributions to $y$ rather than $t$; in both cases, $C_{x_1 y} = 100$.

\subsection{A note on Softmax activation}

Let ${t_1, t_2...t_n}$ represent the output of a softmax transformation on the nodes ${y_1, y_2...y_n}$, such that:
\begin{equation}
A_{t_i} = \frac{e^{A_{y_i}}}{\sum_{i' = 1}^n e^{A_{y_i'}}} 
\end{equation}
Here, $A_{y_1}...A_{y_n}$ are affine functions of their inputs. Let $x$ represent a neuron that is an input to $A_{y_1}...A_{y_n}$, and let $w_{xy_i}$ represent the coefficient of $A_x$ in $A_{y_i}$. Because $A_{y_1}...A_{y_n}$ are followed by a softmax transformation, if $w_{xy_i}$ is the same for all $y_i$ (that is, $x$ contributes equally to all $y_i$), then $x$ effectively has zero contribution to $A_{t_i}$. This can be observed by substituting $A_{y_i} = w_{xy_i}A_x + r_{y_i}$ in the expression for $A_{t_i}$ and canceling out $e^{w_{xy_i}A_x}$ (here, $r_{y_i}$ is the sum of all the remaining terms in the affine expression for $A_{y_i}$)
\begin{equation}
\begin{aligned}
A_{t_i} &= \frac{e^{A_{y_i}}}{\sum_{i' = 1}^n e^{A_{y_i'}}} = \frac{e^{w_{xy_i}A_x + r_{y_i}}}{\sum_{i' = 1}^n e^{w_{xy_{i'}}A_x + r_{y_{i'}}}}\\
           &= \frac{e^{w_{xy_i}A_x + r_{y_i}}}{\sum_{i' = 1}^n e^{w_{xy_i}A_x + r_{y_{i'}}}} = \frac{e^{r_{y_i}}}{\sum_{i' = 1}^n e^{r_{y_{i'}}}} 
\end{aligned}
\end{equation}
As mentioned in the previous subsection, in order to avoid attenuation of signal for highly confident predictions, we should compute $C_{xy_i}$ rather than $C_{xt_i}$. One way to ensure that $C_{xy_i}$ is zero if $w_{xy_i}$ is the same for all $y_i$ is to mean-normalized the weights as follows:
\begin{equation}
\bar{w}_{xy_i} = w_{xy_i} - \frac{1}{n} \sum_{i' = 1}^n w_{xy_{i'}}  
\end{equation} 
This transformation will not affect the output of the softmax, but will ensure that the DeepLIFT scores are zero when a particular node contributes equally to all softmax classes.

\subsection{Weight normalization for constrained inputs}

Let $y$ be a neuron with some subset of inputs $S_y$ that are constrained such that $\sum_{x \in S_y} A_x=c$ (for example, one-hot encoded input satisfies the constraint $\sum_{x \in S_y} A_x = 1$, and a convolutional neuron operating on one-hot encoded rows has one constraint per column that it sees). Let the weights from $x$ to $y$ be denoted $w_{xy}$ and let $b_y$ be the bias of $y$. It is advisable to use normalized weights $\bar{w}_{xy} = w_{xy} - \mu$ and bias $\bar{b}_y = b_y + c\mu$, where $\mu$ is the mean over all $w_{xy}$. We note that this maintains the output of the neural net because, for any constant $\mu$:
\begin{equation}
\begin{aligned}
A_y &= \left(\sum A_x (\bar{w}_{xy} - \mu) \right) + (b_y + c\mu)\\
	&= \left(\sum A_x w_{xy}\right) - \left(\sum A_x \mu\right) + (b_y + c\mu)\\
    &= \left(\sum A_x w_{xy}\right) - c\mu + (b_y + c\mu)\\
    &= \left(\sum A_x w_{xy}\right) + b_y
\end{aligned}
\end{equation}
The normalization is desirable because, for affine functions, the multipliers $m_{xy}$ are equal to the weights $w_{xy}$ and are thus sensitive to $\mu$. To take the example of a convolutional neuron operating on one-hot encoded rows: by mean-normalizing $w_{xy}$ for each column in the filter, one can ensure that the contributions $C_{xy}$ from some columns are not systematically overestimated or underestimated relative to the contributions from other columns.

\vspace{-10px}
\section{Results}

\subsection{Tiny ImageNet}

A model with the VGG16 (Long et al., 2015) architecture was trained using the Keras framework \citep{Chollet2015-ya} on a scaled-down version of the Imagenet dataset, dubbed `Tiny Imagenet'. The images were $64 \times 64$ in dimension and belonged to one of 200 output classes. Results shown in Figure 2; the reference input was an input of all zeros after preprocessing.
\begin{figure}[!ht]
\begin{center}
\includegraphics[scale=0.3]{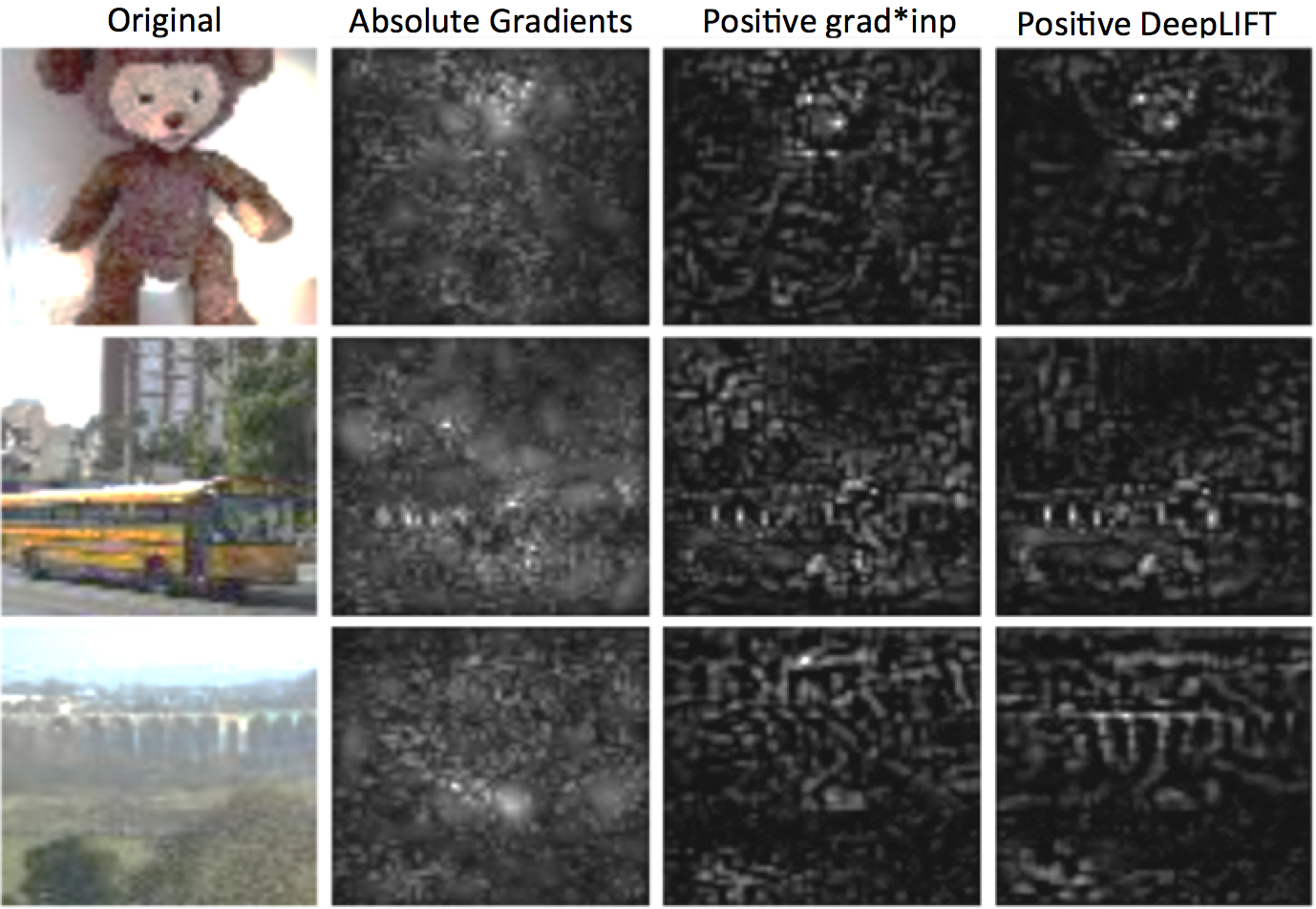}
\caption{Comparison of methods. Importance scores for RGB channels were summed to get per-pixel importance. Left-to-right: original image, absolute value of the gradient (similar to Simonyan et al. which used the two-norm across RGB rather than the sum, and which is related to both Zeiler et al. and Springenberg et al.), positive gradient*input (Taylor approximation, equivalent to Layer-wise Relevance Propagation in Samek et al. but masking negative contributions), and positive DeepLIFT.}
\end{center}
\vspace{-20px}
\end{figure}
\subsection{Genomics}
We apply DeepLIFT to models trained on genomic sequence. The positive class requires that the DNA patterns 'GATA' and 'CAGATG' appear in the length-200 sequence together. The negative class has only one of the two patterns appearing once or twice. Outside the core patterns (which were sampled from a generative model) we randomly sample the four bases A, C, G and T. A CNN was trained using the Keras framework \citep{Chollet2015-ya} on one-hot encoded sequences with 20 convolutional filters of length 15 and stride 1 and a max pool layer of width and stride 50, followed by two fully connected layers of size 200. PReLU nonlinearities were used for the hidden layers. This model performs well with auROC of 0.907. The misclassified examples primarily occur when one of the patterns erroneously arises in the randomly sampled background. % We use dropout with probability 0.3, l1 activity regularization (1e-7) and a weight norm of 7 on the fully connected layers. %We simulate 20K sequences for the positive set and 40K sequences for the negative set split equally between the two patterns.

We then run DeepLIFT to assign an importance score to each base in the correctly predicted sequences. The reference input is an input of all zeros post weight-normalization (see 2.6) of the first convolutional layer (after weight normalization, the linear activation of a convolutional neuron for an input of all zeros is the bias, which is the same as the average activation across all four bases at each position). We compared the results to the gradient*input (Figure 3).

\begin{figure}[!ht]
\vspace{-10px}
\begin{center}
\includegraphics[width=230px,height=30px]{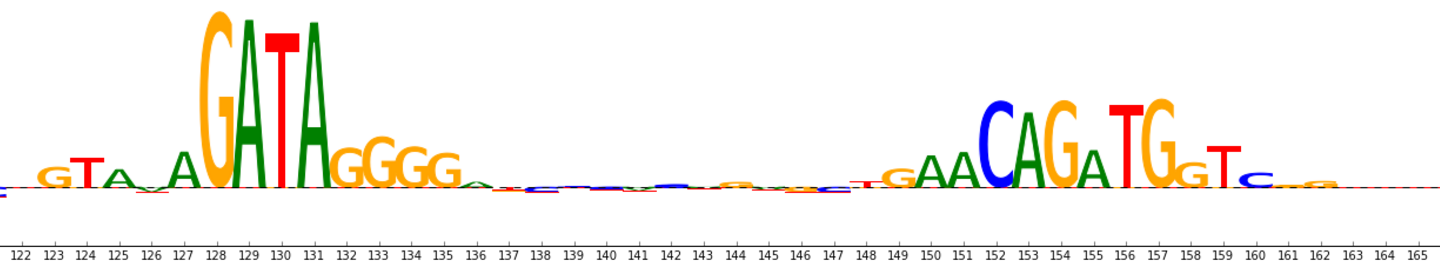}
\includegraphics[width=230px,height=30px]{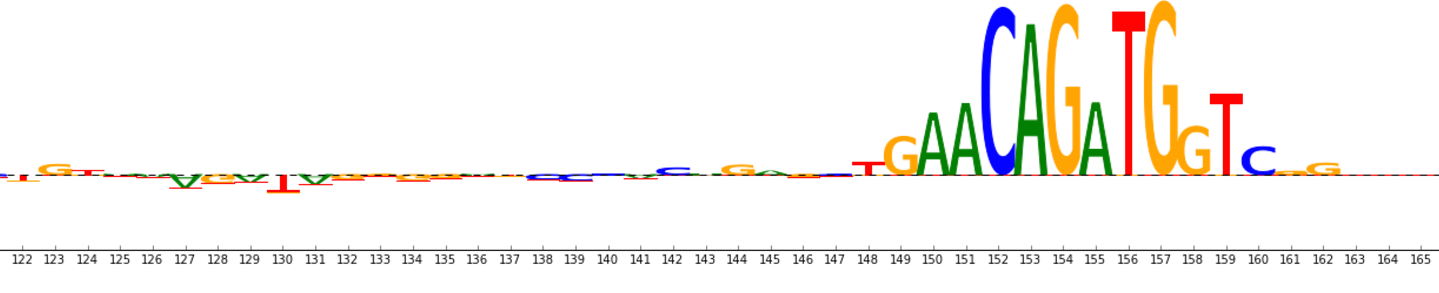}
\caption{DeepLIFT scores (top) and gradient*input (bottom) are plotted for each position in the DNA sequence and colored by the DNA base (due to one-hot encoding, input is either 1 or 0; gradient*input is equivalent to taking the gradient for the letter that is actually present). DeepLIFT discovers both patterns and assigns them large importance scores. Gradient-based methods miss the GATA pattern.}
\end{center}
\vspace{-20px}
\end{figure}

\section{Discussion}

Prevailing feature importance methods such as the saliency maps of Simonyan et al., the deconvolutional nets of Zeiler et al. and the guided backpropagation of Springenberg et al. are variants of computing gradients. As shown in Figure 1, this can give misleading results when the local gradient is zero. DeepLIFT instead considers the deviation from a neuron's reference activity. This makes it capable of handling RNN memory units gated by activations that have vanishing gradients (eg: sigmoid, tanh).

Layer-wise Relevance Propagation (LRP), proposed by Bach et al., does not obviously rely on gradients; however, as we show, if all activations are piecewise linear, LRP reduces to gradient*input (a first-order Taylor approximation of the change in output if the input is set to zero). If all reference activations are zero (as happens when all bias terms are zero and all reference input values are zero), DeepLIFT and LRP give similar results (except that by computing contributions using multipliers, DeepLIFT circumvents the numerical stability problems that LRP faces). In practice, biases are often non-zero, which is why DeepLIFT produces superior results (Figures 2 \& 3).

\subsection{Equivalence of gradient*input to Layer-wise Relevance Propagation}
We show when all activations are piecewise linear and bias terms are included in the calculation, the Layer-wise Relevance Propagation (LRP) of Bach et al., reduces to gradient*input. We refer to Samek et al. (2015) for the concise description of LRP:

{\bf Unpooling}: ``The backwards signal is redirected proportionally onto the location for which the activation was recorded in the forward pass'': This is trivially the same as gradient*input, because the gradient*input will be zero for all locations which do not activation the pooling layer, and equal to the output for the location that does.

{\bf Filtering}: We consider the first rule described in Samek et al., where $z_{ij} = a_i^{(l)} w_{ij}^{(l,l+1)}$ is the weighted activation of neuron $i$ onto neuron $j$ in the next layer, and $l$ is the index of the layer:
\begin{equation}
R_i^{(l)} = \sum_j \frac{z_{ij}}{\sum_i' z_{i'j} + \epsilon \text{ sign}(\sum_{i'} z_{i'j})}R_j^{(l+1)}
\end{equation}

The term involving $\epsilon$ is included to avoid issues of numerical instability when $\sum_i' z_{i'j}$ is near zero. The second rule described in Samek et al. is another variant designed to address the problem of numerical instability. We show that gradient*input gives the exact result as $\epsilon \rightarrow 0$ (i.e. it solves the issue of numerical instability altogether).

Dropping the term for $\epsilon$ and substituting $z_{ij} = a_i^{(l)} w_{ij}^{(l,l+1)}$, we have:
\begin{equation}
R_i^(l) = \sum_j \frac{a_i^{(l)} w_{ij}^{(l,l+1)}}{\sum_i' a_{i'}^{(l)} w_{i'j}^{(l,l+1)}} R_j^{(l+1)}
\end{equation}
Assuming the bias term is included (which would be necessary for the conservation property described in Bach et al. to hold), the denominator is simply the activation of neuron $j$, i.e.:
\begin{equation}
R_i^(l) = \sum_j \frac{a_i^{(l)} w_{ij}^{(l,l+1)}}{a_j^{(l+1)}} R_j^{(l+1)}
\end{equation}
Let us now consider what happens when there are two filtering operations applied sequentially. Let $R_{ik}$ denote the relevance inherited by neuron $i$ in layer $l$ from neuron $k$ in layer $l+2$, passing through the neurons in layer $l+1$. We have:
\begin{equation}
\begin{aligned}
R_{ik}^{(l)} &= \sum_j \frac{a_i^{(l)} w_{ij}^{(l,l+1)}}{a_j^{(l+1)}} \frac{a_j^{(l+1)} w_{jk}^{(l+1,l+2)}}{a_k^{(l+1)}} R_{k}^{(l+2)}\\
              &= \sum_j  \frac{a_i^{(l)} w_{ij}^{(l,l+1)} w_{jk}^{(l+1,l+2)}}{a_k^{(l+1)}} R_{k}^{(l+2)}
\end{aligned}
\end{equation}
Thus, we see that denominator $a_j^{(l+1)}$ for the intermediate layer cancelled out, leaving us with $a_i^{(l)} w_{ij}^{(l,l+1)}w_{jk}^{(l+1,l+2)}$, where $w_{ij}^{(l,l+1)} w_{jk}^{(l+1,l+2)}$ is the gradient of $a_k^{(l+1)}$ with respect to $a_i^{(l)}$. The only term left in the denominator is the activation of the last layer, $a_k^{(l+1)}$; if we set the relevance of neurons in the final layer to be equal to their own activation, then $R_{k}^{(l+2)}$ (assuming $k$ is the last layer) would cancel out $a_k^{(l+1)}$ in the denominator, leaving us with:
\begin{equation}
R_{ik}^{(l)} =  \sum_j  a_i^{(l)} w_{ij}^{(l,l+1)}w_{jk}^{(l+1,l+2)}
\end{equation}
Which is simply equal to the activation $a_i^{(l)}$ multiplied by the gradient of $a_k$ with respect to $a_i^{(l)}$. In situations where the relevance of the last layer is not the same as its activation (which may happen if there is a nonlinear transformation such as a sigmoid, as a sigmoid output of 0.5 occurs when the input is 0), one can simply compute gradient*input with respect to the linear term before the final nonlinearity (which is what we did; for softmax layers, we apply the normalization described in 2.5).

{\bf Nonlinearity}: ``The backward signal is simply propagated onto the lower layer, ignoring the rectification operation'': While this is not obviously the same as gradient*input, it should be noted that when a rectified linear unit is inactive, it has an activation of zero and the rule for filtering (described above) would assign it zero importance. Furthermore, when the rectified linear unit is active, its gradient is 1. Thus, when the unit is inactive, gradient*input is 0 and LRP assigns 0 signal; when a unit is active, gradient*input is equal to the output and LRP assigns all signal. The two approaches converge.

\section{Author contributions}

AS \& PG conceived of DeepLIFT. AS implemented DeepLIFT in software. PG led application to genomics. AYS led application to Tiny Imagenet. AK provided guidance and feedback. AS, PG, AYS \& AK prepared the manuscript.

\bibliographystyle{icml2016}
\bibliography{deeplift}

\end{document}